\documentclass[10pt,twocolumn,letterpaper]{article}

\usepackage{eso-pic}
\usepackage{cvpr}
\usepackage{times}
\usepackage{epsfig}
\usepackage{graphicx}
\usepackage{amsmath}
\usepackage{amssymb}
\usepackage{tikz}
\usepackage{todonotes}
\usepackage{booktabs}       %
\usepackage{paralist}
\usepackage{bm}
\usepackage{caption}

\usepackage{tikz}
\usepackage{pgfplots}
\usetikzlibrary{plotmarks,shapes,snakes,pgfplots.dateplot}
\usetikzlibrary{calc,trees,positioning,arrows,chains,shapes.geometric,%
    decorations.pathreplacing,decorations.pathmorphing,shapes,%
    matrix,shapes.symbols,shadows.blur}
\usepackage{enumerate}

\def\CE{\operatorname{CE}}

\newcommand\lone[1]{\left\lVert #1\right\rVert_1}
\def\LL{{\mathcal L}}
\def\XX{{\mathcal X}}
\def\MM{{\mathcal M}}
\def\CC{{\mathcal C}}
\def\cS{{\mathcal S}}
\def\RR{{\mathbb R}}
\def\II{\mathbf{I}}
\def\bmu{\boldsymbol{\mu}}
\def\bSigma{\boldsymbol{\Sigma}}
\def\bZero{\boldsymbol{0}}
\def\x{\mathbf{x}}
\def\s{\mathbf{s}}
\def\bc{\mathbf{c}}

\def\m{\mathbf{m}}

\def\hx{{\hat{\x}}}
\def\hm{{\hat{\m}}}
\def\hc{{\hat{\bc}}}
\def\hs{{\hat{\s}}}
\def\tx{{\tilde{\x}}}

\def\tc{{\tilde{\bc}}}
\def\ts{{\tilde{\s}}}

\def\xhi{\x_{\mathrm{hi}}}
\def\xlo{\x_{\mathrm{med}}}
\def\hxlo{\hx_{\mathrm{med}}}

\def\ec{E_c}
\def\es{E_s}

\def\lrec{\LL_{\mathrm{rec}}}

\def\genh{G_{\mathrm{enh}}}

\def\ladv{\LL_{\mathrm{adv}}}
\def\lcyc{\LL_{\mathrm{cyc}}}
\def\lseg{\LL_{\mathrm{seg}}}

\def\llc{\LL_{\mathrm{c}}}
\def\lls{\LL_{\mathrm{s}}}
\def\lldist{\LL_{\mathrm{dist}}}

\pgfdeclarelayer{background}
\pgfdeclarelayer{foreground}
\pgfsetlayers{background,main,foreground}

\tikzset{every picture/.style={semithick},every path/.style={very thick,rounded corners,->}}

\tikzset{
  ne/.style={
    draw=none, fill=none,
    font=\sffamily\footnotesize, 
    minimum height=0em,
    text centered},
  nd/.style={
    font=\sffamily\footnotesize, 
    text centered},
  diablo/.style={
    rectangle, 
    rounded corners, 
    draw=black, thick,
    text width=10em, 
    font=\sffamily\footnotesize,
    minimum height=3em, 
    text centered},
  branch/.style ={circle,inner sep=0pt,minimum size=1.5mm,fill=black,draw=black},
  diablo2/.style={
    rectangle, 
    rounded corners, 
    fill=red!10,
    draw=black!80,thick,
    text width=3em, 
    font=\sffamily\footnotesize,
    minimum height=2.0em, 
    text centered},
  dialoss/.style={
    diablo2,
    fill=green!10,
  },
  diablo4/.style={
    diablo2,
    minimum height=1.6em, 
    thin,
  },
  diablo3/.style={
    rectangle, 
    rounded corners, 
    fill=blue!10,
    draw=blue!40,thick,
    text width=3.5em, 
    font=\sffamily\footnotesize\bfseries,
    text=blue,
    minimum height=1.5em,
    text centered},
  line/.style={draw=red,rounded corners,thick, ->, decoration={markings,mark=at position 1 with %
    {\arrow[scale=4,>=stealth]{>}}},postaction={decorate}},
  element/.style={
    tape,
    top color=white,
    bottom color=blue!50!black!60!,
    minimum width=8em,
    draw=blue!40!black!90, very thick,
    text width=10em, 
    minimum height=3.5em, 
    text centered, 
    on chain},
  every join/.style={->,rounded corners,thick,shorten >=1pt},
  decoration={brace},
  lineblue/.style={
    join,line width=.07cm,->,blue!20
  }
}

\tikzset{
  mymatrix/.style={
    nodes={draw,rounded corners=0,thin,minimum size=1em},
    nodes in empty cells,
    column sep=-.5\pgflinewidth,
    row sep=-.5\pgflinewidth,
    blur shadow={shadow blur steps=4, shadow scale=0.9},
    matrix of nodes,
  }
}

\makeatletter
\newcommand{\unmarkedfntext}[1]{{
  \renewcommand{\@makefnmark}{\mbox{}}
  \footnotetext{#1}
}}
\makeatother

\usepackage[pagebackref=true,breaklinks=true,letterpaper=true,colorlinks,bookmarks=false]{hyperref}

\cvprfinalcopy
\def\cvprPaperID{7827} %

\definecolor{amber}{rgb}{1.0, 0.75, 0.2}

\begin{document}

\title{High-Resolution Daytime Translation Without Domain Labels}
\date{}

\author{I. Anokhin$^1$\thanks{Equal contribution.},  \quad
P. Solovev$^1$\footnotemark[1], \quad
D. Korzhenkov$^1$\footnotemark[1], \quad
A. Kharlamov$^1$\footnotemark[1],\\
T. Khakhulin$^{1,3}$, \quad 
A. Silvestrov$^1$, \quad
S. Nikolenko$^{2,1}$, \quad
V. Lempitsky$^{1,3}$, \quad
G. Sterkin$^1$\\
\\
$^1$Samsung AI Center, Moscow \\
$^2$National Research University Higher School of Economics, St.-Petersburg \\
$^3$Skolkovo Institute of Science and Technology, Moscow 
}

\twocolumn[{%
    \renewcommand\twocolumn[1][]{#1}%
    \maketitle
    \centering
    \includegraphics[width=.82\textwidth]{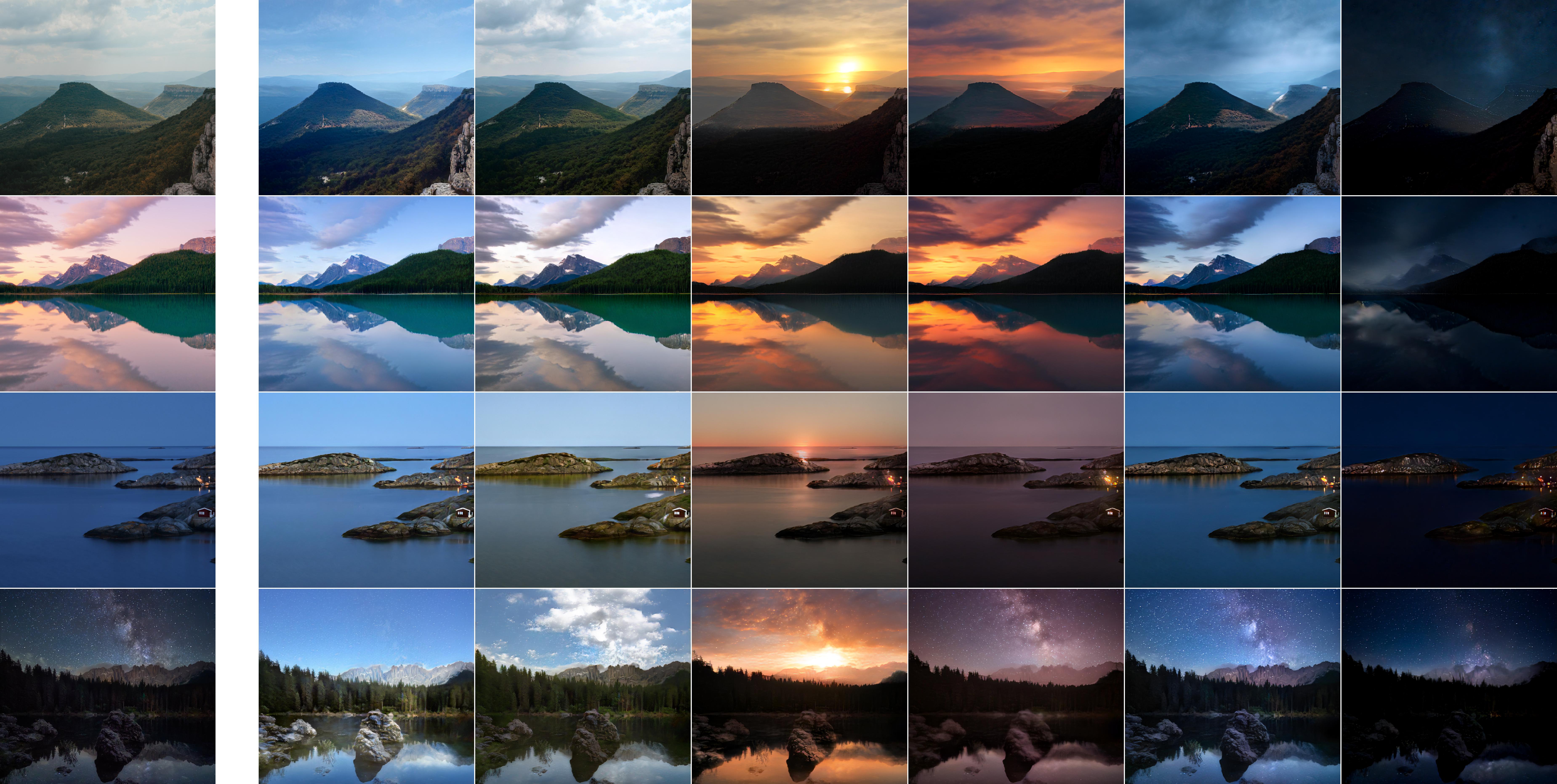}
    \captionof{figure}{Daytime translation results. Left -- original images, right -- translated and enhanced images (one style per column). }
    \label{fig:hook_grid}
    \vspace{.3cm}
}]

\unmarkedfntext{$^\ast$ Equal contribution.}

\begin{abstract}  
Modeling daytime changes in high resolution photographs, e.g., re-rendering the same scene under different illuminations typical for day, night, or dawn, is a challenging image manipulation task. 
We present the high-resolution daytime translation (HiDT) model for this task. HiDT combines a generative image-to-image model and a new upsampling scheme that allows to apply image translation at high resolution.
The model demonstrates competitive results in terms of both commonly used GAN metrics and human evaluation.
Importantly, this good performance comes as a result of training on a dataset of still landscape images with no daytime labels available. 
Our results are available at \url{https://saic-mdal.github.io/HiDT/}.
\end{abstract}

\section{Introduction}
\label{sec:intro}

In this work, we consider the task of generating daytime timelapse videos and pose it as an image-to-image translation problem. Recent image-to-image translation methods have successfully handled the task of conversion between two predefined paired domains~\cite{DBLP:journals/corr/IsolaZZE16,zhu2017unpaired,liu_unsupervised_2017,huang_multimodal_2018} as well as between multiple domains~\cite{choi_stargan_2018,lee_diverse_2018,dritpp,Liu_2019_ICCV}. Given the success of these methods, using image-to-image translation methods to generate daytime changes is a natural idea.

Image-to-image translation approaches require domain labels at training as well as at inference time. The recent FUNIT model~\cite{Liu_2019_ICCV} relaxes this constraint partially. Thus, to extract the style at inference time, it uses several images from the target domain as guidance for translation (known as the \emph{few-shot} setting). The domain annotations are however still needed during training. 

In our task, domains correspond to different times of the day and different lighting, and therefore domain labels are hard to define and hard to solicit from users. Furthermore, while timelapse videos might have provided us with weakly supervised data, we have found that collecting  high-resolution diverse daytime timelapse videos is hard. Therefore, in our work, we aim to develop an image-to-image translation problem suitable for the setting when domain labels are unavailable. 

Thus, as our first contribution, we show how to train a multi-domain image-to-image translation model on a large dataset of unaligned images without domain labels. We demonstrate that the internal bias of the collected dataset, the inductive bias caused by the network architecture, and a specially developed training procedure make it possible to learn style transformations even in this setting. The only external (weak) supervision used by our approach are coarse segmentation maps estimated using an off-the-shelf semantic segmentation network.

As the second contribution, to ensure fine detail preservation, we propose an architecture for image-to-image translation that combines the two well-known ideas: skip connections~\cite{RonnebergerFB15} and adaptive instance normalizations (AdaIN)~\cite{adain}.
We show that such a combination is feasible and leads to an architecture that preserves details much better than currently dominant AdaIN architectures without skip connections.
We evaluate our system against several state-of-the-art baselines through objective measures as well as a user study. While our main focus is the task of photorealistic daytime alteration for landscape images, we also show that such architecture system can be used to handle other multi-domain image stylization/recoloring tasks.
    
Finally, as the third contribution, we address the task of image-to-image translation at high resolution. In our case, as well as in many other settings, training a high-capacity image-to-image translation network directly at high resolution is computationally infeasible. We therefore propose a new enhancement scheme that allows to apply the image-to-image translation network trained at medium resolution for high-resolution images.

The rest of the paper is organized as follows.
Section~\ref{sec:related} reviews related work.
The main Section~\ref{sec:main} presents the High-Resolution Daytime Translation (HiDT) model and the resolution-increasing enhancement model. Section~\ref{sec:eval} presents the results of a comprehensive experimental study, and Section~\ref{sec:concl} concludes the paper.
Representative time-lapse videos generated by our system are provided at the project webpage.

\section{Related work}
\label{sec:related}

\textbf{Unpaired image-to-image translation.} 
The task of image translation aims to transfer images from one domain to another (e.g.\ from summer to winter) or add/remove some image attributes (e.g.\ adding eyeglasses to a portrait).
Many image translation models exploit generative adversarial networks (GAN) with conditional generators to inject information about the target attribute or domain~\cite{choi_stargan_2018}.
Others~\cite{huang_multimodal_2018,dritpp} split input images into content and style representations and subsequently edit the style to obtain the desired effect.
In both cases, most works target the two-domain setting~\cite{zhu2017unpaired} or a setting with several discrete domains~\cite{choi_stargan_2018}.

More closely related to our work, several recent approaches~\cite{huang_multimodal_2018,dritpp,Kotovenko19} split input images into content and style representations and subsequently edit the style to obtain the desired effect.
The most common choice for generators uses adaptive instance normalization (AdaIN) in the encoder-decoder architecture~\cite{adain}.
Providing explicit domain labels is still mandatory for most multi-domain algorithms.
The recently proposed FUNIT model~\cite{Liu_2019_ICCV} is designed for the case when those labels are used only by the conditional discriminator, while the generator is extracting some style code from given samples in the target domain.
In this work, we take the next logical step in the evolution of GAN-based style transfer and do not use domain labels at all.

\textbf{Timelapse generation.} 
The generation of timelapses has attracted some attention from researchers, but most previous approaches use a dataset of timelapse videos for training.
In particular, the work~\cite{shih_data_driven_2013} used a bank of timelapse videos to find the scene most similar to a given image and then exploited the retrieved video as guidance for editing.
Following them, the work~\cite{transient_attributes_2014} used a database of labeled images to create a library of transformations and apply them to image regions similar to input segments.
Both methods rely on global affine transforms in the color space, which are often insufficient to model daytime appearance changes.

Unlike them, a recent paper~\cite{Nam_2019_CVPR} has introduced a neural generation approach.
The authors leveraged two timelapses datasets: one with timestamp labels and another without them, both of different image quality and resolution. 
Finally, a very recent and parallel research~\cite{endo_animating_2019} uses a dataset of diverse videos to solve the daytime appearance change modeling problems. 
Note that the method~\cite{endo_animating_2019} also considers the problem of modeling short-term changes and rapid object motion, which we do not tackle in our  pipeline.
Our approach is different from all previous works for timelapse generation, as it needs neither timestamps nor spatial alignment (such as, e.g.~timelapse frames).

\textbf{High-resolution translation.}
Modern generative models are often hard to scale to high-resolution input images due to memory constraints; most models are trained on either cropped parts or downscaled versions of images. 
Therefore, to generate a plausible image in high resolution one needs an additional enhancement step to upscale the translation output and remove artifacts.
Such enhancement is closely related to the superresolution problem.

The work~\cite{Li_2018_ECCV} compared photorealistic smoothing and image-guided filtering~\cite{he_guided_2013}, and noted that the latter slightly degraded the performance as compared to the former, but led to a significant performance gain.
Another way, proposed in~\cite{Nam_2019_CVPR}, is to apply a different kind of guided upsampling via local color transfer~\cite{progressive_color_transfer_2019}.
However, unlike image-guided filtering, this method does not have a closed-form solution and requires an optimization procedure at inference time.
In~\cite{endo_animating_2019}, the model predicts the parameters of a pixel-wise affine transformation of the downscaled image and then applies bilinear upsampling with these parameters to the full-resolution image. 
Unfortunately, both approaches often produce halo-type artifacts near image edges.

The work most similar to ours in this regard, the \emph{pix2pixHD} model~\cite{DBLP:journals/corr/IsolaZZE16}, developed a separate refinement network. 
Our enhancement model is similar to their approach, as we also use the refinement procedure as a postprocessing step.
But instead of training on the features, we use the output of low-resolution translation directly in a way inspired by classical multi-frame superresolution approaches~\cite{tsai1984multiframe}.

\section{Methods}
\label{sec:main}

The main part of HiDT is an encoder-decoder architecture. The encoder performs decomposition into \textit{style} (vector) and \textit{content} (tensor). The decoder is then able to generate a new image $\hx$ by taking \textit{content} from the content input image $\x$ and \textit{style} from the style input image $\x'$. 

The two components (the content and the style) are combined together using the AdaIN connection~\cite{adain,Liu_2019_ICCV}. The overall architecture has the following structure: the content encoder $\ec$ maps the initial image to a 3D tensor $\bc$ using several convolutional downsampling layers and residual blocks.
The style encoder $\es$ is a fully convolutional network that ends with global pooling and a compressing $1\times 1$ convolutional layer.
The generator $G$ processes $\bc$ with several residual blocks with AdaIN modules inside and then upsamples it.
\begin{figure}[t]
    \centering
    \scalebox{0.9}{
\begin{tikzpicture}[node distance=.3cm]
	\node[diablo2,text width=.5em,fill=blue!10] (x) at (-0.5,0) {$\x$};
	\node[diablo2,text width=.5em,minimum height=16em] (l1) at (.25,0) {};
	\node[diablo2,text width=.5em,minimum height=10em] (l2) at (1.,0) {};
	\node[diablo2,text width=.5em,minimum height=6em] (l3) at (1.75,0) {};
 	\draw (x) -- (l1);
 	\draw (l1) -- (l2);
 	\draw (l2) -- (l3);

 	\node[diablo2,text width=1em,minimum height=3em,fill=blue!10] (c) at (2.5,0) {$\bc$};
 	\draw (l3) -- (c);
 	
 	\node[diablo2,text width=2em,minimum height=2em,fill=blue!10] (s) at (3,1.25) {$\s$};
 	
 	\node[diablo2,text width=2.5em,minimum height=2em,fill=magenta!20] (a1) at (4,0.2) {{AdaIN}};
 	\node[diablo2,text width=2.5em,minimum height=2em,fill=magenta!20] (a2) at (4,-1) {{AdaIN}};
 	\node[diablo2,text width=2.5em,minimum height=2em,fill=magenta!20] (a3) at (4,-2) {{AdaIN}};
	
	\draw (c) -- (a1.200);
	\draw (l2.east |- a2.200) -- (a2.200);
	\draw (l1.east |- a3.200) -- (a3.200);

	\draw (s.south) |- (a1.160);
	\draw (s.south) |- (a2.160);
	\draw (s.south) |- (a3.160);
	
    \node[diablo2,text width=.5em,minimum height=4em] (d0) at (5.,0) {};
	\node[diablo2,text width=.5em,minimum height=10em] (d1) at (5.75,0) {};
	\node[diablo2,text width=.5em,minimum height=16em] (d2) at (6.5,0) {};
    \node[diablo2,text width=.75em,fill=blue!10] (hx) at (7.25,0) {$\hx$};

    \draw (a1) -- (a1.east -| d0.west);
    \draw (d0) -- (d1);
    \draw (d1) -- (d2);
    \draw (a2) -- (a2.east -| d1.west);
    \draw (a3) -- (a3.east -| d2.west);

    \draw (d2) -- (hx);

\end{tikzpicture}
}
    \caption{Diagram of the Adaptive U-Net architecture: an encoder-decoder network with dense skip-connections and content-style decomposition~$(\bc, \s)$.}
    \label{fig:adaptive_unet}
\end{figure}
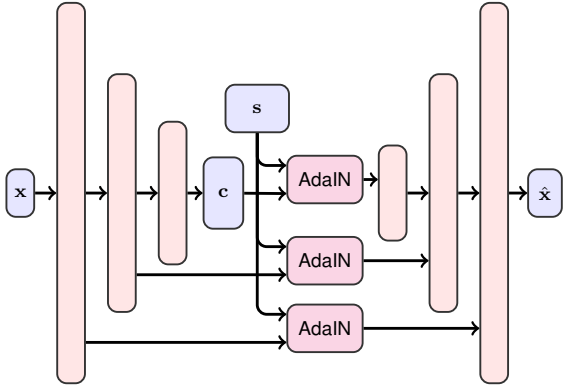

To create a plausible daytime landscape image, the model should preserve fine details from the original image.  
To satisfy this requirement, we enhance the encoder-decoder architecture with skip connections between the downsampling part of tje encoder $\ec$ and the upsampling part of the generator $G$.
Regular skip connections would also ``leak'' the style of the initial input into the output.
Therefore, we introduce an additional convolutional block with AdaIN~\cite{adain} and apply it to the skip connections (see Fig.~\ref{fig:adaptive_unet}).

\subsection{Learning}

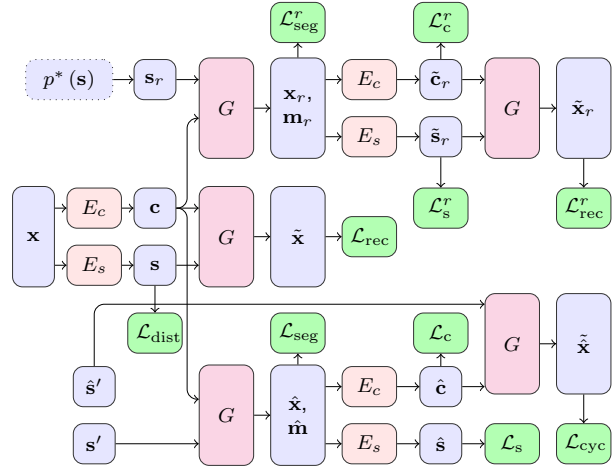
\begin{figure}[t]
    \centering
    \colorlet{colorloss}{green!30}

\def\widf{1.5em}
\def\widx{1em}
\def\widxx{1.5em}
\def\widl{1.8em}
\def\heix{4em}
\def\heixx{8em}

\def\heione{2.25}
\def\heitwo{3.6}
\def\heithree{4.4}
\def\heitwothree{4}

\def\heionetop{2.65}
\def\heionebot{1.85}

\def\heibottom{8.3}
\def\heibottop{8.7}
\def\heibotbot{7.9}

\def\heiy{9.5}

\def\heifive{5.0}
\def\heifivetop{5.15}
\def\heifivebot{4.5}

\def\heifour{6.5}
\def\heifourtop{6.9}
\def\heifourbot{6.1}

\def\heiz{0.3}
\def\heiztop{0.7}
\def\heizbot{-0.1}

\def\xinputimage{1.3}
\def\xone{2.15}
\def\xtwo{3.0}
\def\xthree{4.}
\def\xfour{5.}
\def\xfive{6.}
\def\xsix{7.}
\def\xseven{8.}
\def\xeight{9.}

\def\contentencoder{$\ec$}
\def\styleencoder{$\es$}
\def\decoder{$G$}

\def\xa{\x_{a}}
\def\xb{\x_{b}}
\def\ca{\bc_{a}}
\def\cb{\bc_{b}}
\def\sa{\s_{a}}
\def\sb{\s_{b}}
\def\txa{\tilde{\x}_{a}}
\def\txb{\tilde{\x}_{b}}
\def\tma{\tilde{\m}_{a}}
\def\tmb{\tilde{\m}_{b}}
\def\xab{\x_{ab}}
\def\xba{\x_{ba}}
\def\mab{\m_{ab}}
\def\mba{\m_{ba}}

\scalebox{0.95}{
\begin{tikzpicture}
	\node[diablo4,text width=\widx,minimum height=\heix,fill=blue!10] (x) at (\xinputimage,\heifour) {$\x$};

	\node[diablo4,text width=\widf] (ec1) at (\xone,\heifourtop) {\contentencoder};
	\node[diablo4,text width=\widf] (es1) at (\xone,\heifourbot) {\styleencoder};
	\node[diablo4,text width=\widx,fill=blue!10] (cx) at (\xtwo,\heifourtop) {$\bc$};
	\node[diablo4,text width=\widx,fill=blue!10] (sx) at (\xtwo,\heifourbot) {$\s$};
	\node[diablo4,text width=\widx,fill=blue!10] (sxht) at (\xone,\heithree) {$\hs'$};
	\draw[thin] (x.east |- ec1.west) -- (ec1);
	\draw[thin] (x.east |- es1.west) -- (es1);
	\draw[thin] (ec1) -- (cx);
	\draw[thin] (es1) -- (sx);

	\node[diablo4,text width=\widx,fill=blue!10] (sy) at (\xone,\heitwo) {$\s'$};
	\node[diablo4,dotted,text width=2.8em,fill=blue!10] (rand) at (1.8,\heibottop) {$p^{\ast}\left(\s\right)$};
	\node[diablo4,text width=\widx,fill=blue!10] (sr) at (\xtwo,\heibottop) {$\s_r$};
	\draw[thin] (rand) -- (sr);

	\node[diablo4,text width=\widf,minimum height=\heix,fill=magenta!20] (g1) at (\xthree,\heifour) {\decoder};
	\draw[thin] (cx) -- (cx.east -| g1.west);
	\draw[thin] (sx) -- (sx.east -| g1.west);

	\node[diablo4,text width=\widf,minimum height=\heix,fill=magenta!20] (g12) at (\xthree,\heitwothree) {\decoder};
	\draw[thin] (cx.east) -- ++(.15,0) |- (g12.150);
	\draw[thin] (sy) -- (sy.east -| g12.west);

	\node[diablo4,text width=\widf,minimum height=\heix,fill=magenta!20] (g3) at (\xthree,\heibottom) {\decoder};
	\draw[thin] (cx.east) -- ++(.15,0) |- (g3.200);
	\draw[thin] (sr) -- (sr.east -| g3.west);

	\node[diablo4,text width=\widxx,minimum height=\heix,fill=blue!10] (tx) at (\xfour,\heifour) {$\tx$};
	\draw[thin] (g1) -- (tx);

	\node[diablo4,text width=\widxx,minimum height=\heix,fill=blue!10] (xab) at (\xfour,\heitwothree) {$\hx$, $\hm$};
	\draw[thin] (g12) -- (xab);

	\node[diablo4,text width=\widxx,minimum height=\heix,fill=blue!10] (rx) at (\xfour,\heibottom) {$\x_r$, $\m_r$};
	\draw[thin] (g3) -- (rx);

	\node[diablo4,text width=\widf] (ec2) at (\xfive,\heithree) {\contentencoder};
	\node[diablo4,text width=\widf] (es2) at (\xfive,\heitwo) {\styleencoder};
	\draw[thin] (xab.east |- ec2.west) -- (ec2);
	\draw[thin] (xab.east |- es2.west) -- (es2);

	\node[diablo4,text width=\widx,fill=blue!10] (tc) at (\xsix,\heithree) {$\hc$};
	\node[diablo4,text width=\widx,fill=blue!10] (ts) at (\xsix,\heitwo) {$\hs$};
	\draw[thin] (ec2) -- (tc);
	\draw[thin] (es2) -- (ts);

	\node[diablo4,text width=\widf] (ec3) at (\xfive,\heibottop) {\contentencoder};
	\node[diablo4,text width=\widf] (es3) at (\xfive,\heibotbot) {\styleencoder};
	\draw[thin] (rx.east |- ec3.west) -- (ec3);
	\draw[thin] (rx.east |- es3.west) -- (es3);

	\node[diablo4,text width=\widx,fill=blue!10] (rc) at (\xsix,\heibottop) {$\tc_r$};
	\node[diablo4,text width=\widx,fill=blue!10] (rs) at (\xsix,\heibotbot) {$\ts_r$};
	\draw[thin] (ec3) -- (rc);
	\draw[thin] (es3) -- (rs);

	\node[diablo4,text width=\widf,minimum height=\heix,fill=magenta!20] (g4) at (\xseven,\heibottom) {\decoder};
	\draw[thin] (rc) -- (rc.east -| g4.west);
	\draw[thin] (rs) -- (rs.east -| g4.west);

	\node[diablo4,text width=\widf,minimum height=\heix,fill=blue!10] (rrx) at (\xeight,\heibottom) {$\tx_r$};
	\draw[thin] (g4) -- (rrx);

	\node[diablo4,text width=\widf,minimum height=\heix,fill=magenta!20] (g5) at (\xseven,\heifive) {\decoder};
	\draw[thin] (tc) -- (tc.east -| g5.west);
	\draw[thin] (sxht.north) |- (g5.125); 
	
	\node[diablo4,text width=\widf,fill=colorloss] (srec) at (\xseven,\heitwo) {$\lls$};
	\draw[thin] (ts) -- (srec);

	\node[diablo4,text width=\widf,minimum height=\heix,fill=blue!10] (srx) at (\xeight,\heifive) {$\tilde{\hx}$};
	\draw[thin] (g5) -- (srx);

	\node[diablo4,text width=\widf,fill=colorloss] (xarec) at (\xfive,\heifour) {$\lrec$};
	\draw[thin] (tx.east |- xarec.west) -- (xarec);

    \node[diablo4,text width=\widf,fill=colorloss] (lrecr) at (\xeight,\heifourtop) {$\lrec^r$};
	\draw[thin] (rrx) -- (lrecr);
    
    \node[diablo4,text width=\widf,fill=colorloss] (lrecs) at (\xeight,\heitwo) {$\lcyc$};
	\draw[thin] (srx) -- (lrecs);
    
    \node[diablo4,text width=\widf,fill=colorloss] (lseg) at (\xfour,\heifivetop) {$\lseg$};
	\draw[thin] (xab) -- (lseg);

    \node[diablo4,text width=\widf,fill=colorloss] (lc) at (\xsix,\heifivetop) {$\llc$};
	\draw[thin] (tc) -- (lc);

    \node[diablo4,text width=\widf,fill=colorloss] (lsegr) at (\xfour,\heiy) {$\lseg^r$};
	\draw[thin] (rx) -- (lsegr);

    \node[diablo4,text width=\widf,fill=colorloss] (lcr) at (\xsix,\heiy) {$\llc^r$};
	\draw[thin] (rc) -- (lcr);

    \node[diablo4,text width=\widf,fill=colorloss] (lsr) at (\xsix,\heifourtop) {$\lls^r$};
	\draw[thin] (rs) -- (lsr);

    \node[diablo4,text width=\widf,fill=colorloss] (ldist) at (\xtwo,\heifivetop) {$\lldist$};
	\draw[thin] (sx.south) -- (ldist.north);
\end{tikzpicture}
}
    \caption{HiDT learning data flow. We show half of the (symmetric) architecture; $\s'=\es(\x')$ is the style extracted from the other image $\x'$, and $\hs'$ is obtained similarly to $\hs$ with $\x$ and $\x'$ swapped.
    Light blue nodes denote data elements; light green, loss functions; others, functions (subnetworks).
    Functions with identical labels have shared weights. Adversarial losses are omitted for clarity.
    }
    \label{fig:prgan}
    \vspace{-10pt} %
\end{figure}

Overall, the architecture is trained using a reconstruction loss as well as a number of additional losses (Fig.~\ref{fig:prgan}). During training, the decoder predicts not only the input image $\x$ but also its semantic segmentation mask $\m$ (produced by a pretrained network~\cite{Sun_2019_CVPR}).
While we do not aim to achieve state-of-the-art segmentation as a by-product, having the segmentation loss helps to control the style transfer and to preserve the semantic layout.
Importantly, segmentation masks are \emph{not} given as input to the networks, and are thus not needed at inference time.

\textbf{Notation.} Denote the space of input images by $\XX$, their segmentation masks by $\MM$, and individual images with segmentation masks by $\x,\m\in\XX\times\MM$. Denote the space of latent content codes $\bc$ is $\bc\in\CC$, and the space of latent style codes $\s$ is $\s\in\cS$ (as we will see below, $\cS=\RR^3$ while $\CC$ has a more complex structure).
To extract $\bc$ and $\s$ from an image $\x$, HiDT employs two encoders:
  $\ec:\XX\to\CC$ extracts the content representation $\bc$ of the input image $\x$, and 
  $\es:\XX\to\cS$ extracts the style representation $\s$ of the input image $\x$.
Given a content code $\bc\in\CC$ and a style code $\s\in\cS$,
the decoder (generator) $G:\CC\times\cS\to\XX\times\MM$ produces a new image $\hx$ and the corresponding segmentation mask $\hm$. In particular, one can combine content from $\x$ and style from a different image $\x'$ as
    $\left( \hx, \hm \right) = G\left(\ec(\x), \es(\x')\right).$
We call the result of the combination the \textit{translated image} (and the \textit{translated mask}). 

Also, during training we consider random style codes $\s_r$ sampled from a prior distribution $p^*$ on $\cS$. Then we get a \textit{random style image} (and a \textit{random style mask}) by applying the decoder to the content code $\bc$ and the random style $\s_r$ respectively. During learning for each batch, we take the reconstructed images/masks, the translated images/masks (where the images are paired, and the styles are swapped) and the random style images/masks.

\textbf{Image reconstruction loss.} The image reconstruction loss $\lrec$ is defined as the $L_1$-norm of the difference between original and reconstructed images.
It is applied in three different ways: (1) to the reconstruction $\tx$ of the original image $\x$, $\lrec = \lone{\tx - \x}$, (2) to the reconstruction $\tx_r$ of the random style image $\x_r$, $\lrec^r = \lone{\tx_r - \x_r}$, and (3) to the reconstruction $\tilde{\hx}$ of the  image $\x$ obtained from the content of the stylized image $\hx$ and the style of the stylized image $\hx'$ (cross cycle consistency): $\lcyc = \lone{\tilde{\hx} - \x}$, where $\tilde{\hx}=G(\hc, \hs')$ (see Fig.~\ref{fig:prgan}).

\textbf{Segmentation loss.} The segmentation loss $\lseg$ is used together with the image reconstruction loss and is defined as the cross entropy $\CE(\m,\hm)=-\sum_{(i,j)}m_{i,j}\log \hat{m}_{i,j}$ between the original $\m$ and reconstructed $\hm$ segmentation masks.
It is applied in two ways: first, to the translated mask $\hm$ , $\lseg = \CE(\m, \hm)$, and then to the random style mask $\m_r$: $\lseg^r = \CE(\m,\m_r)$.

\textbf{Adversarial loss.} We use two discriminators, namely, the unconditional discriminator and the conditional discriminator, where the style vector is used as conditioning. Both discriminators consider translated and random style images as fakes. Both discriminators are trained with the least squares GAN approach~\cite{mao2017least}. We utilize the projection conditioning scheme~\cite{miyato_cgans_2018} and detach the styles from the computational graph when feeding them to the conditional discriminator during the generator update step.

\textbf{Latent reconstruction losses.} We enforce cycle consistency with respect to the style and the content codes. We pass the translated and the random style images into the encoders, and compute the losses between the resulting style (content) and the style (content) that the respective translated or the random style image was obtained from. We apply the $L_1$ loss to content codes as well as to the style codes.

\textbf{Style distribution loss.} To enforce the structure of the space of extracted style codes, the style distribution loss inspired by the CORAL approach~\cite{sun_correlation_2017}, is applied to a pool of styles collected from a number of previous training iterations.
Namely, for a given pool size $T$ we collect the styles $\{ \s^{\left(1\right)}, \dots, \s^{\left(T\right)} \}$ from past minibatches with the \emph{stop gradient} operation applied. We then add styles $\s$ and $\s'$ (which are part of the current computational graph) to this pool, and calculate the mean vector $\hat{\bmu}_s$ and covariance matrix $\hat{\bSigma}_s$ using the updated pool.
Then the style distribution loss  matches empirical moments of the resulting distribution to the moments of the prior distribution $\mathcal{N} \left(\bZero, \II\right)$:
$ \lldist = \lone{\hat{\bmu}_T} + \lone{\hat{\bSigma}_T - \II} + \lone{diag(\hat{\bSigma}_T) - \mathbf{1}}. $
Since the space $\cS = \RR^3$ is low-dimensional, and our target is the unit normal distribution $\mathcal{N} \left(\bZero, \II\right)$, this simplified approach suffices to enforce the structure in the space of latent codes.
After computing the loss value, the oldest styles are removed from the pool to keep its size at $T$.

\textbf{Total loss function.} Thus, overall HiDT is jointly training the style encoder, content encoder, generator, and discriminator with the following objective:
\begin{multline*}
    \min\limits_{E_c, E_s, G} \max\limits_{D} \; \LL\left(E_c, E_s, G, D \right) = \lambda_1(\ladv + \ladv^r) + \\
    + \lambda_2(\lrec + \lrec^r + \lcyc) + 
    \lambda_3(\lseg + \lseg^r) + \\
    +  \lambda_4(\llc + \llc^r) +\lambda_5\lls + \lambda_{6}\lls^r + \lambda_{7}\lldist.
\end{multline*}
Hyperparameters $\lambda_1,\ldots,\lambda_{7}$ define the relative importance of the components in the overall loss function; they have been chosen by hand and will be shown below.

During our experiments, we have observed that the projection discriminator significantly improves the results, while removing the segmentation loss function sometimes leads to undesirable hallucinations in the generator (see Fig.~\ref{fig:no_segm_failure} for an example). 
However, the model is still well trained without segmentation loss function and gets a comparable user preference score.
We provide a further ablation study in the supplementary material.

\subsection{Enhancement postprocessing}
\label{sec:enhance}

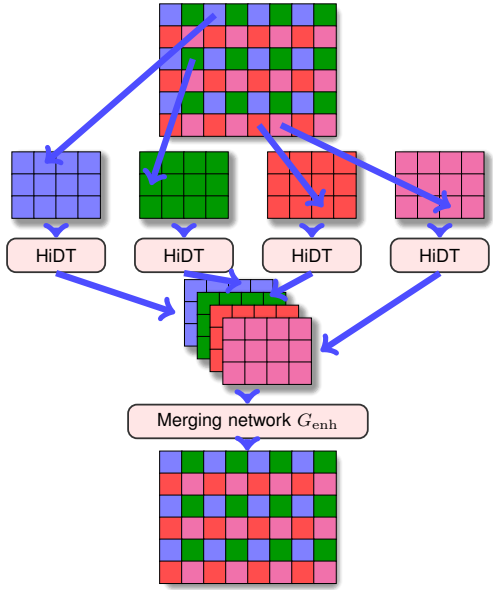
\begin{figure}[t]\centering
    \scalebox{.85}{
        \colorlet{mc1}{blue!50}
\colorlet{mc2}{green!60!black}
\colorlet{mc3}{red!70}
\colorlet{mc4}{magenta!70}

\def\sxone{-3}
\def\sxtwo{-1}
\def\sxthree{1}
\def\sxfour{3}
\def\sxfive{10.4}
\def\syone{7.7}
\def\sytwo{6.6}
\def\sythree{5.4}
\def\syfour{4.0}
\def\syfive{2.5}

\begin{tikzpicture}

\matrix[mymatrix] (M) at (0,9.5) {
|[fill=mc1]| & |[fill=mc2]| & |[fill=mc1]| & |[fill=mc2]| & |[fill=mc1]| & |[fill=mc2]| & |[fill=mc1]| & |[fill=mc2]| \\
|[fill=mc3]| & |[fill=mc4]| & |[fill=mc3]| & |[fill=mc4]| & |[fill=mc3]| & |[fill=mc4]| & |[fill=mc3]| & |[fill=mc4]| \\
|[fill=mc1]| & |[fill=mc2]| & |[fill=mc1]| & |[fill=mc2]| & |[fill=mc1]| & |[fill=mc2]| & |[fill=mc1]| & |[fill=mc2]| \\
|[fill=mc3]| & |[fill=mc4]| & |[fill=mc3]| & |[fill=mc4]| & |[fill=mc3]| & |[fill=mc4]| & |[fill=mc3]| & |[fill=mc4]| \\
|[fill=mc1]| & |[fill=mc2]| & |[fill=mc1]| & |[fill=mc2]| & |[fill=mc1]| & |[fill=mc2]| & |[fill=mc1]| & |[fill=mc2]| \\
|[fill=mc3]| & |[fill=mc4]| & |[fill=mc3]| & |[fill=mc4]| & |[fill=mc3]| & |[fill=mc4]| & |[fill=mc3]| & |[fill=mc4]| \\
};

\matrix[mymatrix,nodes={fill=mc1}] (M1) at (\sxone,\syone) {
& & & \\
& & & \\
& & & \\
};

\matrix[mymatrix,nodes={fill=mc2}] (M2) at (\sxtwo,\syone) {
& & & \\
& & & \\
& & & \\
};

\matrix[mymatrix,nodes={fill=mc3}] (M3) at (\sxthree,\syone) {
& & & \\
& & & \\
& & & \\
};

\matrix[mymatrix,nodes={fill=mc4}] (M4) at (\sxfour,\syone) {
& & & \\
& & & \\
& & & \\
};

\draw[line width=1mm, blue!70] (M-1-3.center) -- (M1-1-2.center);
\draw[line width=1mm, blue!70] (M-3-2.center) -- (M2-2-1.center);
\draw[line width=1mm, blue!70] (M-6-5.center) -- (M3-3-3.center);
\draw[line width=1mm, blue!70] (M-6-6.center) -- (M4-3-3.center);

\node[diablo2,minimum height=1.5em,text width=3.7em] (t1) at (\sxone,\sytwo) {HiDT};
\node[diablo2,minimum height=1.5em,text width=3.7em] (t2) at (\sxtwo,\sytwo) {HiDT};
\node[diablo2,minimum height=1.5em,text width=3.7em] (t3) at (\sxthree,\sytwo) {HiDT};
\node[diablo2,minimum height=1.5em,text width=3.7em] (t4) at (\sxfour,\sytwo) {HiDT};

\draw[line width=1mm, blue!70] (M1) -- (t1);
\draw[line width=1mm, blue!70] (M2) -- (t2);
\draw[line width=1mm, blue!70] (M3) -- (t3);
\draw[line width=1mm, blue!70] (M4) -- (t4);

\matrix[mymatrix,nodes={fill=mc1}] (MM1) at (-.3,\sythree+.3) {
& & & \\
& & & \\
& & & \\
};

\matrix[mymatrix,nodes={fill=mc2}] (MM2) at (-.1,\sythree+.1) {
& & & \\
& & & \\
& & & \\
};

\matrix[mymatrix,nodes={fill=mc3}] (MM3) at (.1,\sythree-.1) {
& & & \\
& & & \\
& & & \\
};

\matrix[mymatrix,nodes={fill=mc4}] (MM4) at (.3,\sythree-.3) {
& & & \\
& & & \\
& & & \\
};

\node[diablo2,minimum height=1.5em,text width=10em] (mnet) at (0,\syfour) {Merging network $\genh$};

\matrix[mymatrix] (MM) at (0,\syfive) {
|[fill=mc1]| & |[fill=mc2]| & |[fill=mc1]| & |[fill=mc2]| & |[fill=mc1]| & |[fill=mc2]| & |[fill=mc1]| & |[fill=mc2]| \\
|[fill=mc3]| & |[fill=mc4]| & |[fill=mc3]| & |[fill=mc4]| & |[fill=mc3]| & |[fill=mc4]| & |[fill=mc3]| & |[fill=mc4]| \\
|[fill=mc1]| & |[fill=mc2]| & |[fill=mc1]| & |[fill=mc2]| & |[fill=mc1]| & |[fill=mc2]| & |[fill=mc1]| & |[fill=mc2]| \\
|[fill=mc3]| & |[fill=mc4]| & |[fill=mc3]| & |[fill=mc4]| & |[fill=mc3]| & |[fill=mc4]| & |[fill=mc3]| & |[fill=mc4]| \\
|[fill=mc1]| & |[fill=mc2]| & |[fill=mc1]| & |[fill=mc2]| & |[fill=mc1]| & |[fill=mc2]| & |[fill=mc1]| & |[fill=mc2]| \\
|[fill=mc3]| & |[fill=mc4]| & |[fill=mc3]| & |[fill=mc4]| & |[fill=mc3]| & |[fill=mc4]| & |[fill=mc3]| & |[fill=mc4]| \\
};

\draw[line width=1mm, blue!70] (t1.south) -- (MM1.west);
\draw[line width=1mm, blue!70] (t2.south) -- (MM2.north);
\draw[line width=1mm, blue!70] (t3.south) -- (MM3.70);
\draw[line width=1mm, blue!70] (t4.south) -- (MM4.east);

\draw[line width=1mm, blue!70] (MM4.245) -- (mnet);
\draw[line width=1mm, blue!70] (mnet) -- (MM-1-5.135);

\end{tikzpicture}
    }
    \caption{Enhancement scheme: the input is split into subimages (color-coded) that are translated individually by HiDT at medium resolution. The outputs are then merged using the merging network $\genh$. For illustration purposes, we show upsampling by a factor of two, but in the experiments we use a factor of four. We also apply bilinear downsampling (with shifts -- see text for detail) rather than strided subsampling when decomposing the input into medium resolution images.
    }
    \label{fig:shuffled}
\end{figure}

Training image-to-image translation on high resolution images is infeasible due to both memory and computation time constraints. In principle, our architecture can be trained at medium resolution and applied to high resolution images in a fully convolutional way. Alternatively,  guided filtering~\cite{he_guided_2013} can be used to upsample results of processing at medium resolution. Although both of these techniques show good results in most cases, they have limitations. A fully convolutional application might yield scene corruption due to limited receptive field, which is the case with sunsets where multiple suns might be drawn, or water reflections where the border between sky and water surface might be confused.
Guided filtering, on the other hand, works great with water or sun but fails if small details like twigs were changed by the style transfer procedure. It also often generates halo artefacts near the horizon and other high-contrast borders. Finally, we have found that a  superresolution architecture \cite{wang_esrgan:_2019} does not generalize well even to well-looking translated images, effectively amplifying translation artefacts. 

Inspired by existing multiframe image restoration methods~\cite{tsai1984multiframe}, we propose to apply translation multiple times at medium resolution and then use a separate merging network $\genh$ to combine the results into a high-resolution translated image. More specifically, we consider a high resolution image $\xhi$ (in our experiments, $1024\times 1024$). We then consider sixteen shifted versions of $\xhi$ denoted as $\{\xhi^{(i)}\}_i$, each having the same size as $\xhi$ and obtained with integer displacement spanning the range $[0;4]$ in $x$ and $y$ (missing pixels are filled with zeros). The shifted images are then downsampled bilinearly resulting in sixteen medium-resolution images $\{\xlo^{(i)}\}_i$, from which the original image $\xhi$ can be easily recovered. 

We then apply HiDT to each of the medium-resolution images separately, getting translated medium-resolution images $\{ \hxlo^{(i)} \}_i$, $\hxlo^{(i)}= G(\ec(\xlo^{(i)}), \es(\xlo^{(i)}))$.
These frames are stacked into a single tensor in a fixed order and are fed to the merging network $\genh$ that outputs the translated high-resolution image. The process is illustrated in Fig.~\ref{fig:shuffled}.

The merging network $\genh$ is trained in a semi-supervised mode on two datasets: paired and unpaired. To obtain a paired dataset, we use HiDT in an ``autoencoder mode'' (i.e.~without changing the style). To obtain each training pair, we take a high-res image, decompose it into sixteen medium-resolution images, and pass them through the HiDT architecture without changing the style. For the unpaired dataset collection we use the same procedure, but the style is being sampled from normal distribution (since we used it as a prior during training). The merging network is thus shown stacks of resulting images and is tasked with restoring the original image. At test time, we can use a new style $\s'$, when translating each of the medium-resolution images. The output of the merging network will then correspond to the high-resolution input image $\xhi$ translated to the style $\s'$.

We note the similarity of our approach to~\cite{pix2pixhd}, with the difference being that we use several RGB images as input instead of feature maps. During training, we use the same losses as \emph{pix2pixHD}~\cite{pix2pixhd}, namely perceptual, feature matching, and adversarial loss functions. We apply only adversarial loss for the unpaired data. %

\section{Experiments}
\label{sec:eval}

\begin{table}[t]
    \centering
    {\small
    \setlength{\tabcolsep}{8pt}
    \begin{tabular}{c|l|ccc}
    \toprule
    $N$ & \begin{tabular}[l]{@{}l@{}} HiDT vs   \\  \texttt{method} \end{tabular}  &\begin{tabular}[r]{@{}l@{}}User $\uparrow$ \\ score\end{tabular} & p-value &  \multicolumn{1}{c}{\begin{tabular}[c]{@{}c@{}}Adjusted \\ p-value\end{tabular}}  \\
     \hline
    1 &  \begin{tabular}[l]{@{}l@{}}\texttt{DRIT}\end{tabular}    & 0.53    &  0.997 & 1.0 \\
     &  \begin{tabular}[l]{@{}l@{}}\texttt{FUNIT-T}\end{tabular}  & 0.51    &  0.904 &  0.999  \\
      &  \begin{tabular}[l]{@{}l@{}}\texttt{FUNIT-O}\end{tabular}   & 0.57 & 0.999  &  1.0 \\
      \hline
    5 &  \begin{tabular}[l]{@{}l@{}}\texttt{FUNIT-T}\end{tabular}  & 0.48  & 0.024   &  0.179 \\
     &  \begin{tabular}[l]{@{}l@{}}\texttt{FUNIT-O}\end{tabular}   & 0.55  & 0.481 &   1.0 \\
      \hline
    10 & \texttt{FUNIT-T}  & 0.47  & 0.001  &  0.011   \\
     &  \begin{tabular}[r]{@{}l@{}}\texttt{FUNIT-O} \end{tabular}  & 0.57  & 0.999 & 1.0    \\
    
    \bottomrule
    \end{tabular}
    }
    \captionof{table}{User preference study of HiDT against the baselines. 
    $N$ is the number of styles averaged in the few-shot setting. 
    The user score is the share of users that choose HiDT in the pairwise comparison. Our results show that all methods are competitive.
    The increase of $N$ leads to the better quality of FUNIT-T.
    }
    \label{tab:i2i_human_preferences}
    \vspace{-10pt}
\end{table}

\begin{table}[t]
    \small
    \centering
    {\small
    \setlength{\tabcolsep}{5pt}
    \begin{tabular}{l|lllll}
    \toprule
    Method & \begin{tabular}[c]{@{}l@{}}DIPD$\downarrow$\\ swapped\end{tabular}  & \begin{tabular}[c]{@{}l@{}}DIPD$\downarrow$\\ random\end{tabular} & CIS$\uparrow$   & \multicolumn{1}{c}{\begin{tabular}[c]{@{}c@{}}IS$\uparrow$\\ random\end{tabular}} & \multicolumn{1}{c}{\begin{tabular}[c]{@{}c@{}}IS$\uparrow$\\ swapped\end{tabular}} \\

    \hline
    \texttt{FUNIT-T}  & 1.168                                                  &   -                                                     & 1.535 & -                                                                       & 1.615       \\                                                 
    \texttt{DRIT} & 0.863                                                  & 1.018                                                 & 1.203 & 1.251                                                                   & 1.577 \\
    \texttt{HiDT-AE} & 0.321  &  -  & 1.179 &  - & 1.524 \\
    \texttt{HiDT}   & 0.691                                                  & 0.88                                                 & 1.559 & 1.673                                                                   & 1.605 \\
    
    \bottomrule
                                      
    \end{tabular}
    }
    \captionof{table}{Performance comparison of three models using a hold-out dataset.
    FUNIT is not applicable in the random setting.
    According to the selected metrics, none of the models shows complete superiority over the others. }
    \label{tab:i2i_metrics}
\end{table}

\begin{figure}[ht!]
    \centering
    \setlength{\tabcolsep}{2pt}
    \begin{tabular}{cc}
    \includegraphics[width=.49\linewidth]{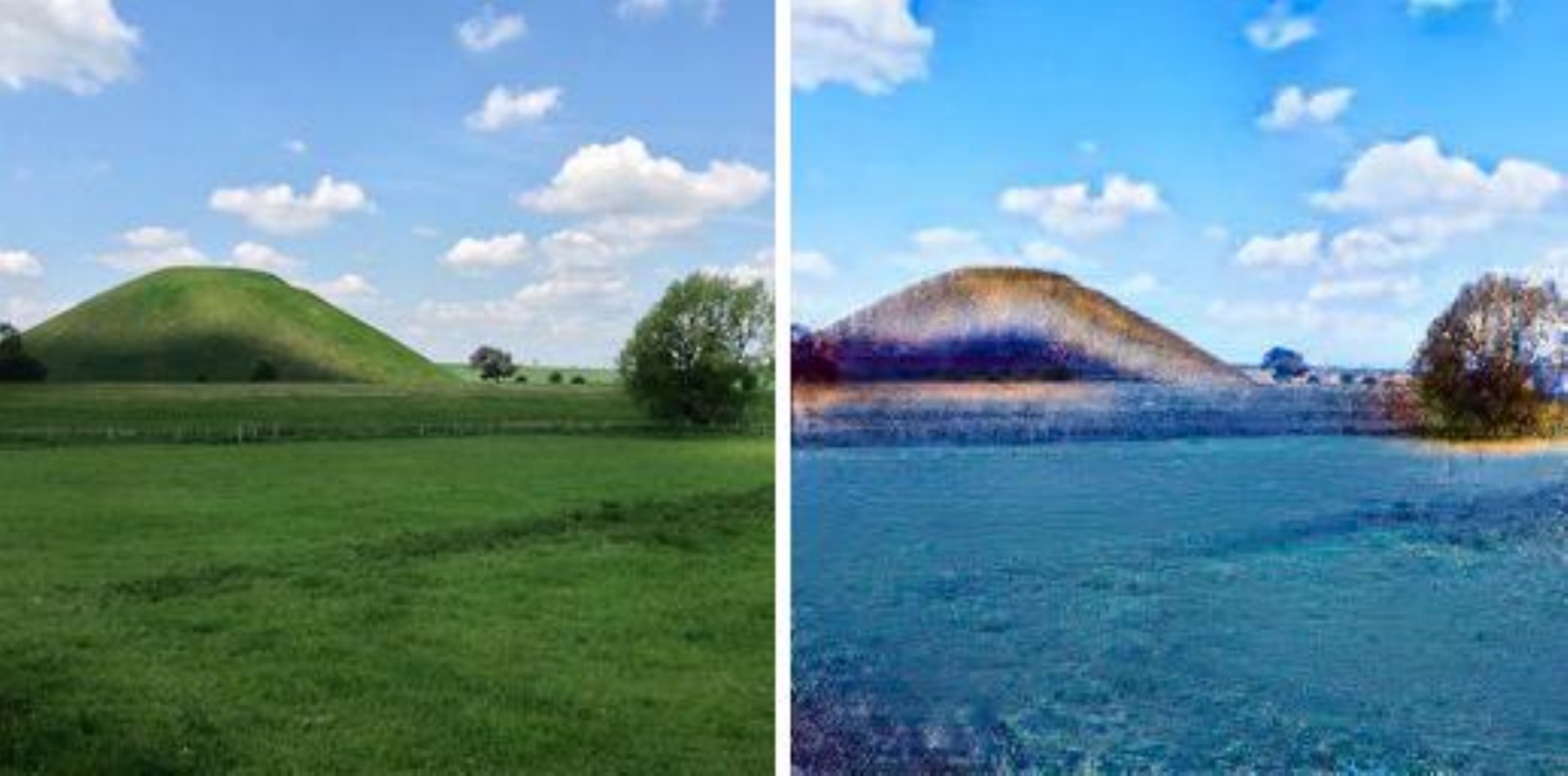} &
    \includegraphics[width=.49\linewidth]{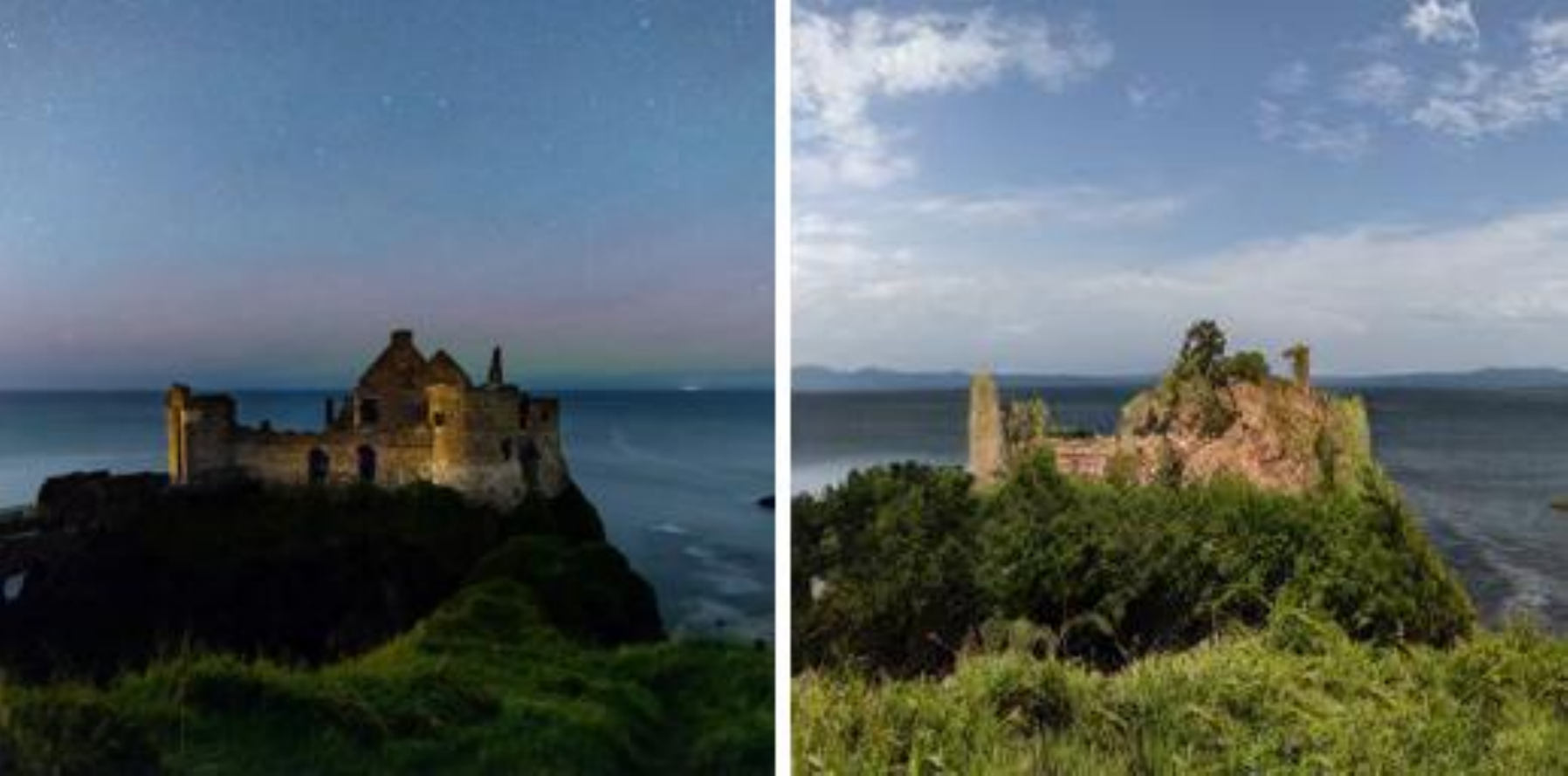}
    \\ \footnotesize (a) & 
    \footnotesize  (b) 
    \end{tabular}
    \caption{ Training without segmentation losses is prone to failures of semantic consistency.  {Left}: original images. {Right:} transferred images. (a) Our ablated model, trained without auxiliary segmentation task, turns grass into water; (b) FUNIT hallucinates grass on the building.}
    \label{fig:no_segm_failure}
    \vspace{-15pt}
\end{figure}

\subsection{Daytime translation}

\textbf{Training details.}
In our experiments, the content encoder has two downsampling and four residual blocks; after each downsampling, only five channels are used for skip connections in order to limit the information flow through them.
The style encoder contains four downsampling blocks. The output of the style encoder is a three-channel tensor, which is averaged-pooled into a three-dimensional vector.
The decoder has five residual blocks with AdaIN layers and two upsampling blocks.
AdaIN parameters are computed from the style vector via three-layer fully-connected network.
Both discriminators are multi-scale, with three downsampling levels.
We trained the translation model for 450 thousand iterations with batch size four on a single NVIDIA Tesla P40. For training, the images were downscaled to the resolution of $256\times 256$.
The loss weights were set to $\lambda_1 = 5, \lambda_2 = 2, \lambda_3 = 3, \lambda_4 = 1, \lambda_5 = 0.1, \lambda_6 = 4, \lambda_7 = 1$.
We used the Adam optimizer~\cite{Kingma14} with $\beta_1 = 0.5$, $\beta_2 = 0.999$, and initial learning rate $0.0001$ for both generators and discriminators, halving the learning rate every $200000$ iterations. 

\textbf{Dataset and daytime classifier.}
Following previous works, we collected a dataset of 20{,}000 landscape photos from the Internet.
A small part of these images were manually labeled into four classes (night, sunset/sunrise, morning/evening, noon) using a crowdsourcing platform.
A ResNet-based classifier was trained on those labels and applied to the rest of the dataset.
We used predicted labels in two ways:
\begin{inparaenum}[(1)]
    \item to balance the training set for image translation models with respect to daytime classes; 
    \item to provide domain labels for baseline models.
\end{inparaenum}
Segmentation masks were produced by an external state of the art model~\cite{Sun_2019_CVPR} and reduced to nine classes: sky, grass, ground, mountains, water, buildings, trees, roads, and humans.

\textbf{Baselines.}
We used two recent image-to-image translation models as baselines: FUNIT~\cite{Liu_2019_ICCV} and Multi-domain DRIT++~\cite{dritpp} (refered to as DRIT for brevity).
Both of them use domain labels.
We trained the models on our dataset: DRIT with original hyperparameters, and FUNIT with both original (FUNIT-O) and properly tuned (FUNIT-T) hyperparameters.
At inference time, FUNIT transfers the original image using styles extracted from other images, while DRIT in addition can transfer to randomly sampled styles.
As another weak baseline, we train our model with only the autoencoding loss $\lrec$ (HiDT-AE).
The trained HiDT-AE still produces some color shifting when the styles are swapped; the result does not resemble the target daytime well enough, although it preserves the content (details) well.

\textbf{Evaluation metrics.}
To compare our model with the baselines, we use several metrics, also commonly employed in previous works.
The \emph{domain-invariant perceptual distance} (DIPD)~\cite{huang_multimodal_2018,Liu_2019_ICCV} is the $L_2$ distance between normalized \emph{Conv5} features of the original image and its translated version.
It is used to measure content preservation.
The \emph{Inception score} (IS)~\cite{inception_score} assesses the photorealism of generated images.
We  use  the  classifier  described  above  to predict the domain label of the translated image. 
Styles for the translation may be either sampled from the prior distribution $p^{\ast}(\s)$ (IS-random) or extracted from other images (IS-swapped). 
The \emph{conditional inception score} (CIS)~\cite{huang_multimodal_2018} measures the diversity of translation results, which is suitable for our multi-domain setting.
We calculate CIS for style swapping translation.
To estimate the visual plausibility and photorealism of translation results, we use \emph{human evaluation} with the following protocol.
The assessors on a crowd-sourcing platform\footnote{\url{https://toloka.yandex.ru/}} were shown triplets containing
\begin{inparaenum}
    \item the original image,
    \item the image translated with our method, and 
    \item the image translated using one of the baseline models.
\end{inparaenum}
We also show assessors the target label (time of day) and ask to choose the image that looks better with respect to both details preserved from the original image and the correct time of day.
As both our model and FUNIT support the few-shot setting, styles for translation were obtained by averaging $N$ styles extracted from images with the corresponding labels ($N=1,5,10$).
Assessment time was limited to two minutes per task, and original images were independently collected from the Internet.
For each compared pair of methods, we generated 500 triplets, and each triplet was assessed by five different workers. 

\begin{figure}[t]
    \centering
    \includegraphics[width=\columnwidth]{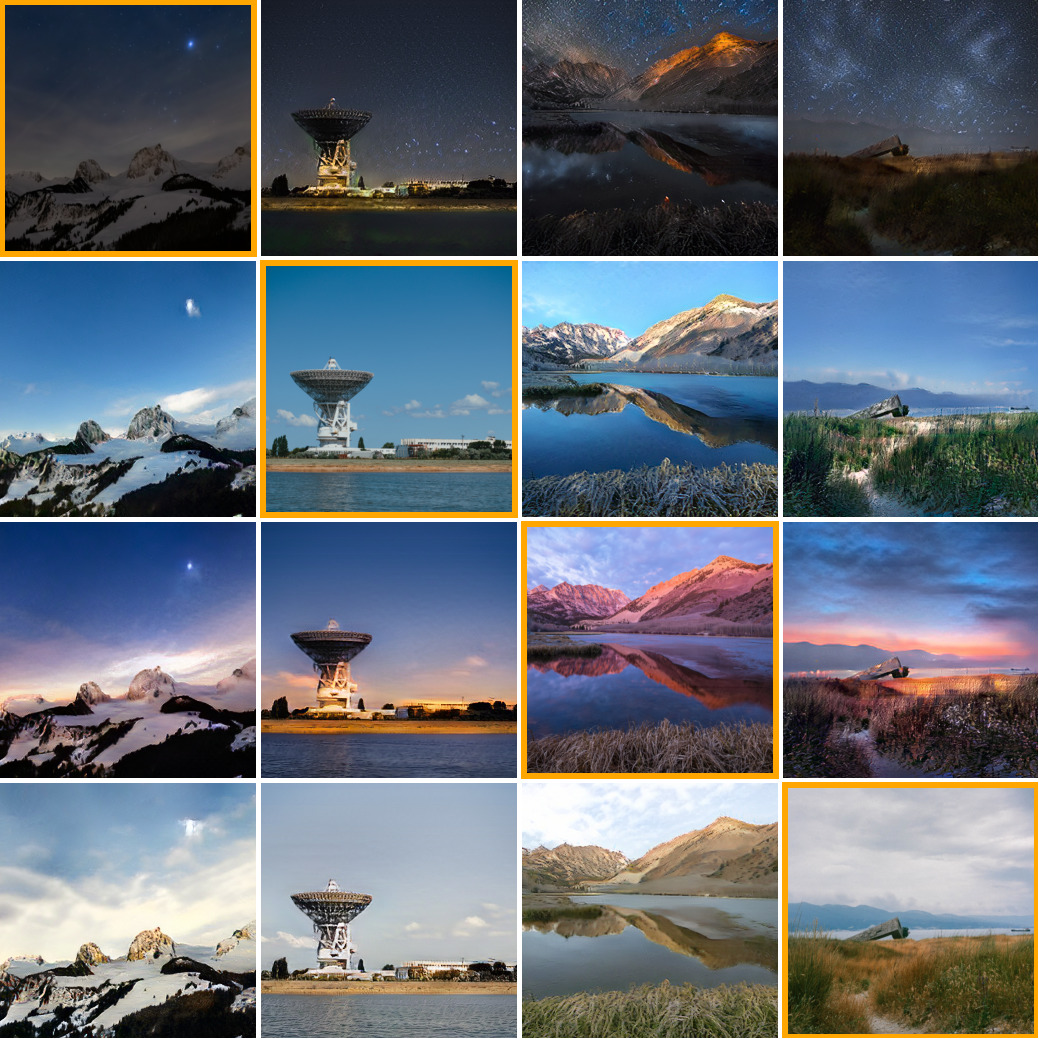}
    \caption{Swapping styles between two images. Original images are shown on the main diagonal. The examples show that HiDT is capable to swap the styles between two real images while preserving details. }
    \label{fig:swapped_grid}
    \vspace{-10pt}
\end{figure}

\begin{figure}[t]
    \centering
    \includegraphics[width=\columnwidth]{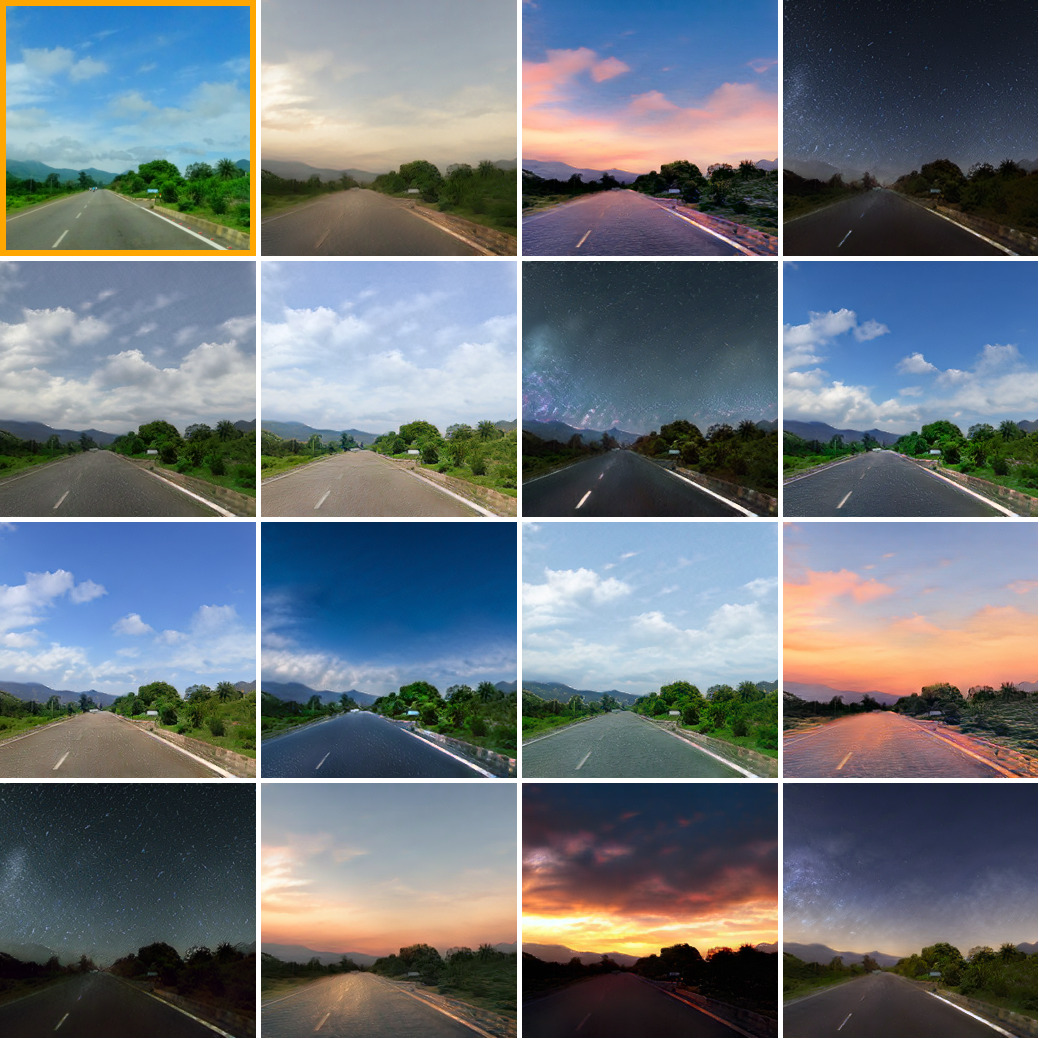}
    \caption{ The original content image (top left), transferred to randomly sampled styles from prior distribution. The results demonstrate the diversity of possible outputs. }
    \label{fig:random_grid}
    \vspace{-10pt}
\end{figure}

\begin{figure}[ht!]
    \centering
    \includegraphics[width=\columnwidth]{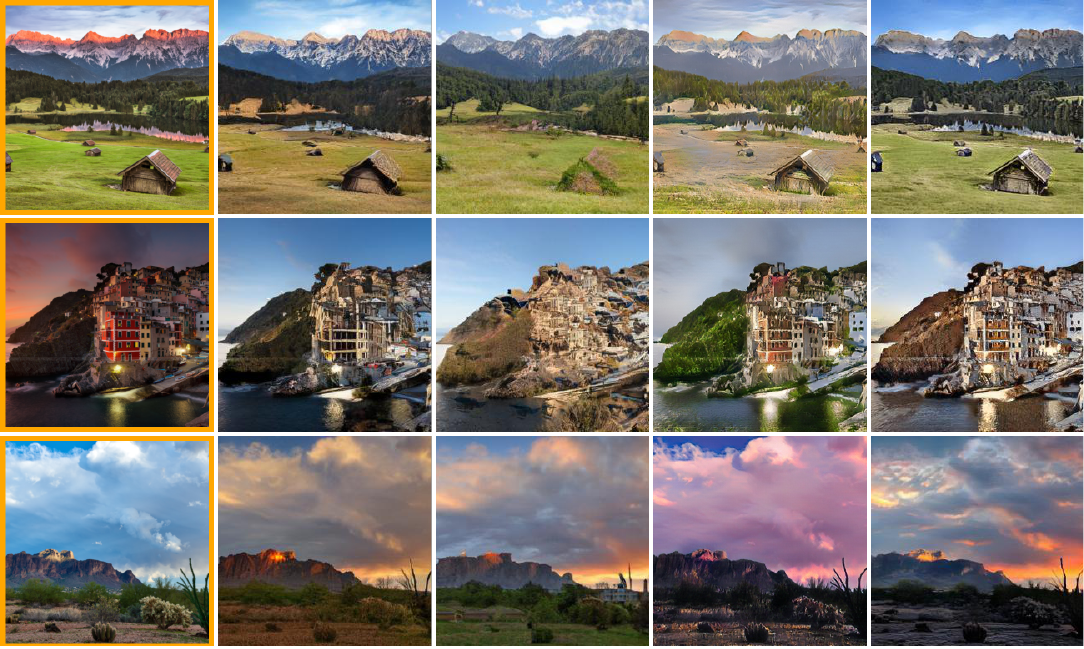}
    \caption{Comparison with baselines. Columns, left to right: the original image, FUNIT-T, FUNIT-O, DRIT, HiDT (ours).
    Our model, trained and applied without knowledge about domain labels, has translation quality similar to the models that require such supervision. }
    \label{fig:competitors}
    \vspace{.3cm}

    \centering
    \includegraphics[width=\columnwidth]{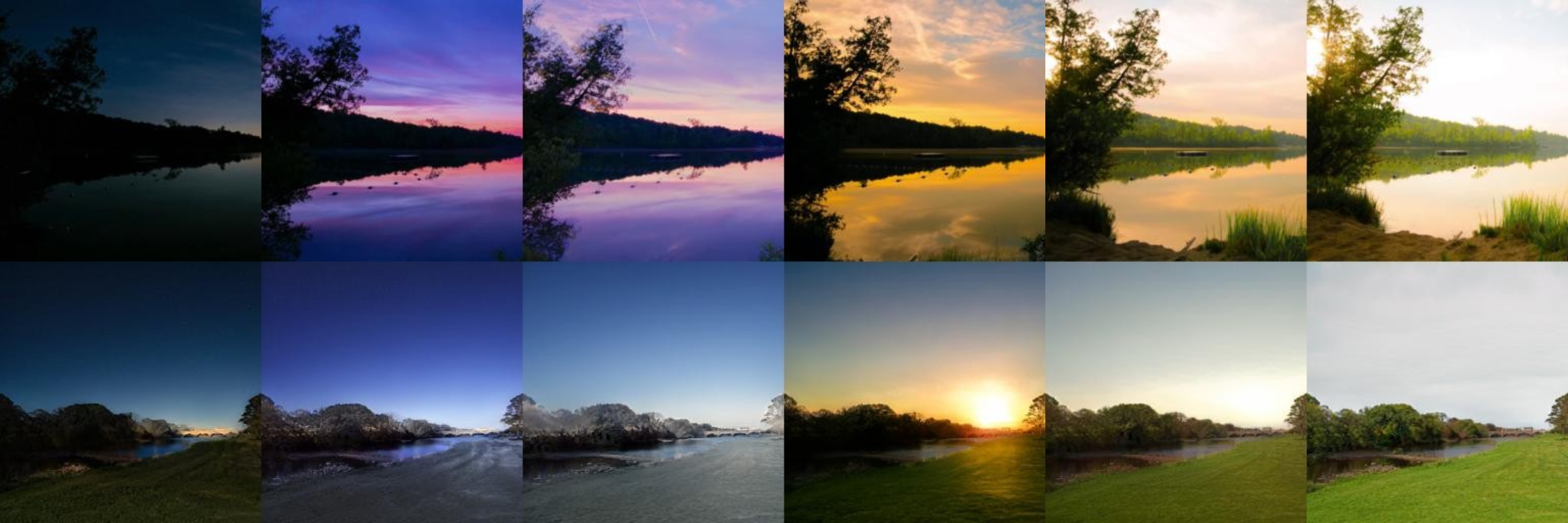}
    \caption{Timelapse generation using styles extracted from a real video.
    {Top}: frames from a guidance video.
    {Bottom}: timelapse generated from a single image using extracted styles.}
    \label{fig:timelapse_result}
\end{figure}

\textbf{Results.}
Sample results of our image translation model are shown in the teaser figure on the first page.
Fig.~\ref{fig:swapped_grid} shows style swapping between different images, while image translation with styles randomly sampled from the prior distribution is shown in Fig.~\ref{fig:random_grid}.
In these experiments, we applied the truncation trick known for improving the average output quality~\cite{brock2018large,Karras_2019_CVPR} at inference time.
Random styles are sampled with reduced variance, and the styles extracted from other images are interpolated with the style extracted from the original image.
One important application of our model is daytime timelapse generation using some video as a guidance; we showcase frames from such a timelapse in Fig.~\ref{fig:timelapse_result}. 

A qualitative comparison of our model with baselines is shown in Fig.~\ref{fig:competitors}.
Results of different models are hard to distinguish, which is supported by our human evaluation study (Table~\ref{tab:i2i_human_preferences}).
We report user preference of our model over the baselines and evaluate its statistical significance, applying the one-tailed binomial test to the hypothesis ``User score equals 0.5'' against ``User score is less than 0.5''.
Due to multiple hypothesis testing, we also apply the Holm-Sidak adjustment and show adjusted p-values.
Table~\ref{tab:i2i_human_preferences} suggests that unlabeled training is sufficient for time-of-day translation.
Traditional image-to-image translation metrics are summarized in Table~\ref{tab:i2i_metrics}.
Again, all models are basically on par with each other, with different winners according to different metrics. 

\begin{figure}[ht!]
        \centering
    \includegraphics[width=\columnwidth]{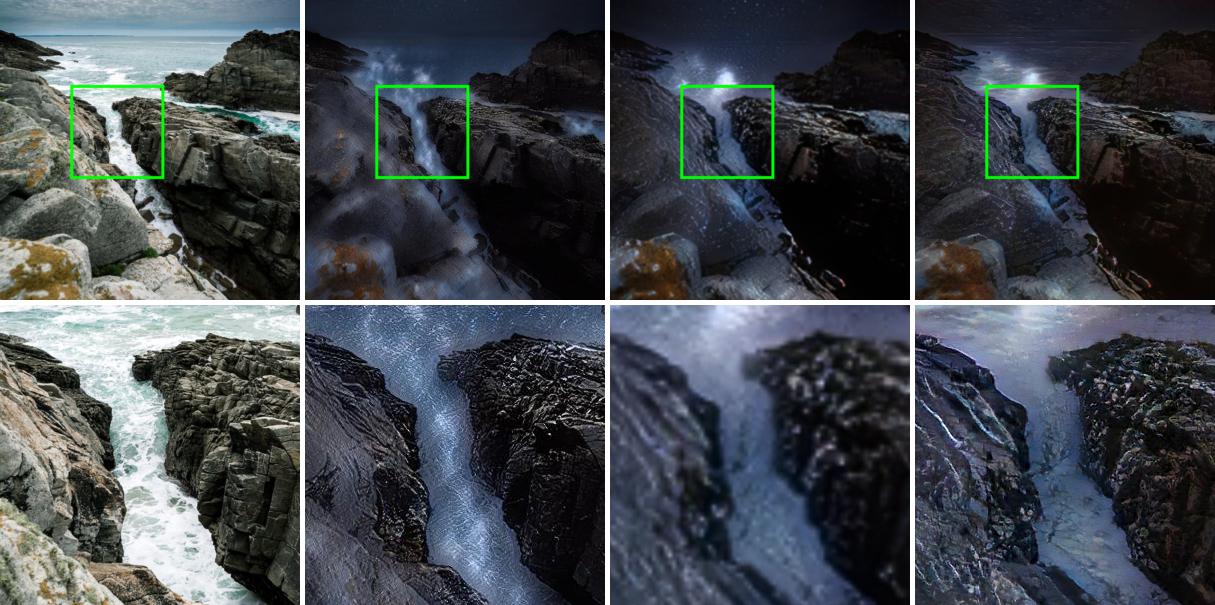}
    \caption{Enhancement of our translation network outputs with different methods.
    Columns, left to right: original image; result of our translation network applied directly to the hi-res input; low-res translation output upsampled with Lanczos' method; the result of our enhancement scheme.
    In this example, direct fully-convolutional application to hi-res turns water into sky with stars, while the enhancement network preserves the semantics of the scene.}
    \label{fig:enhance}
        \vspace{.3cm}
        \centering

    \includegraphics[width=\columnwidth]{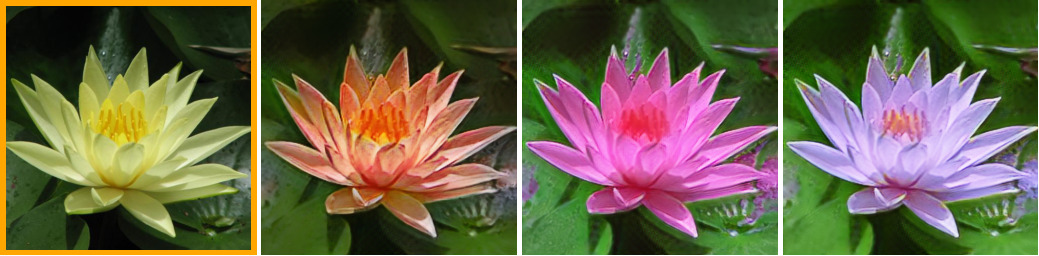}
    \caption{A flower image (left) translated to several randomly sampled styles by HiDT trained on Oxford Flowers dataset.}
        \label{fig:flower_60k}
\end{figure}

\begin{figure}
    \centering
    \includegraphics[width=\columnwidth]{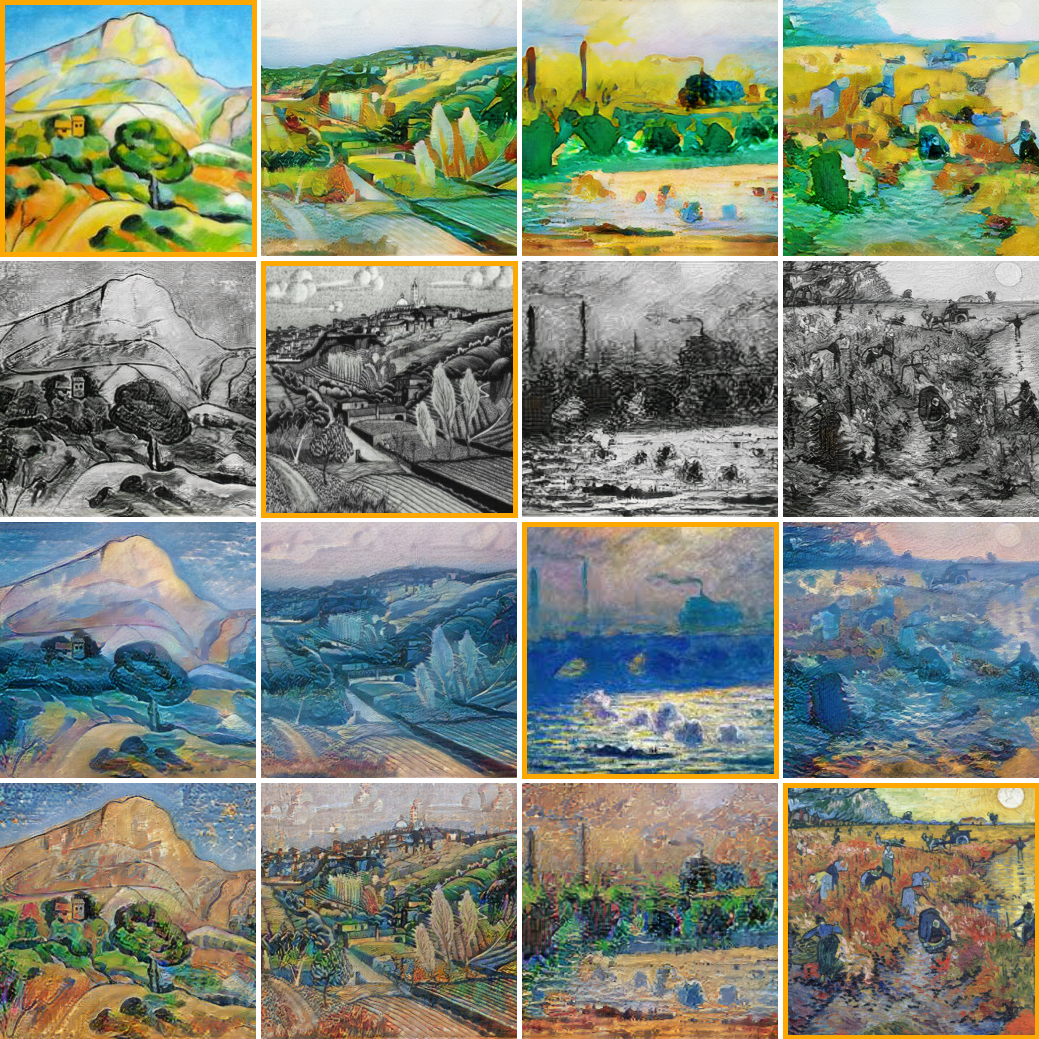}
    \caption{Style swapping for the HiDT system trained on a paintings dataset. The main diagonal contains original paintings and off-diagonal entries correspond to translated results. Plausbile translations obtained by HiDT in this case, suggests its generality. }
        \label{fig:artists}

\end{figure}
\subsection{High-resolution translation}

\textbf{Training details.}
For the merging network, we used the RRBDNet architecture from ESRGAN~\cite{wang_esrgan:_2019} with five residual blocks for $\genh$ and a multiscale discriminator with three scales and five layers. We used multiplier coefficients of 10 for perceptual and feature matching losses and the unit weight for adversarial loss. We set learning rate of $0.0001$ for both the merging network and the discriminator.

\textbf{Baselines.}
We compare the proposed enhancement scheme with the following baselines:
\begin{inparaenum}[(1)]
    \item fully convolutional application of the translation network to a high-resolution image, 
    \item Lanczos upsampling.
\end{inparaenum}
The \emph{pix2pixHD}~\cite{pix2pixhd} enhancement scheme requires supervision for translated images.
Therefore, we do not use \emph{pix2pixHD} as a baseline.

\textbf{Results.}
The resulting downsampled images produced with the enhancement procedure are presented in Fig.~\ref{fig:hook_grid}, and a detailed example is shown in Fig.~\ref{fig:enhance}.
The latter figure contains image patch produced by different models and shows that our model is more plausible than the result of direct Lanczos upsampling: the rightmost patch contains more details from the original.

\subsection{Additional task}

To show the generality of the proposed HiDT approach, we additionally trained the image translation model on the \emph{Flowers} dataset~\cite{Nilsback08} for 60{,}000 iterations. Segmentation masks and associated losses were not used in this experiment. The results of translation to random styles (with no enhancement) are presented in Fig.~\ref{fig:flower_60k}.  We have also applied HiDT to the WikiArt dataset of paintings (for which we have increased the dimensionality of the style space to $12$). The result of style swapping in this case is shown in Fig.~\ref{fig:artists}.

\section{Conclusion}
\label{sec:concl}

We have presented an image-to-image translation model that does not rely on domain labels during either training or inference. The new enhancement scheme shows promising results for increasing the resolution of translation outputs. We have shown that our model is able to learn daytime translation for high-resolution landscape images and provided qualitative evidence that our approach can be generalized to other domains.

The results show that our method is on par with state of the art baselines that require labels at least at training time. Our model can generate images using styles extracted from images, as well as sampled from the prior distribution. An appealing straightforward application of our model is the generation of timelapses from a single image (the task currently mainly tackled with paired datasets). One direction for further work would be to unite the translation and enhancement networks into a single model trained end-to-end.

\newpage
{\small
\bibliographystyle{ieee_fullname}
\bibliography{egbib}
}

\end{document}